\pdfoutput=1

\documentclass[11pt]{article}

\usepackage[]{acl}

\usepackage{times}
\usepackage{latexsym}

\usepackage[T1]{fontenc}

\usepackage[utf8]{inputenc}

\usepackage{microtype}

\usepackage{microtype}
\usepackage{graphicx}
\usepackage{subfigure}
\usepackage{booktabs} 

\usepackage{multirow}
\usepackage{makecell}
\usepackage{amsmath}
\usepackage{setspace}

\usepackage{color,xcolor}
\usepackage{amssymb}
\usepackage{algorithm} 
\usepackage{algorithmic}
\usepackage{CJKutf8}

\title{Categorizing Semantic Representations for Neural Machine Translation}

\author{
 Yongjing Yin$^{1,2}\thanks{\ \  This work was done as an intern at Pattern Recognition Center, WeChat AI, Tencent Inc, China.}$, Yafu Li$^{1,2}$, Fandong Meng$^{3}$, Jie Zhou$^{3}$, Yue Zhang$^{2,4}$ \\
 $^1$ Zhejiang University\\
 $^2$ School of Engineering, Westlake University\\
 $^3$ Pattern Recognition Center, WeChat AI, Tencent Inc\\
 $^4$ Institute of Advanced Technology, Westlake Institute for Advanced Study\\ 
 \quad{\{yinyongjing,liyafu\}@westlake.edu.cn} \quad{\{fandongmeng,withtomzhou\}@tencent.com} \\
 \quad{yue.zhang@wias.org.cn} \\
}

\begin{document}
\maketitle
\begin{abstract}
Modern neural machine translation (NMT) models have achieved competitive performance in standard benchmarks. However, they have recently been shown to suffer limitation in compositional generalization, failing to effectively learn the translation of atoms (e.g., words) and their semantic composition (e.g., modification) from seen compounds (e.g., phrases), and thus suffering from significantly weakened translation performance on unseen compounds during inference.
We address this issue by introducing categorization to the source contextualized representations.
The main idea is to enhance generalization by reducing sparsity and overfitting, which is achieved by finding prototypes of token representations over the training set and integrating their embeddings into the source encoding. 
Experiments on a dedicated MT dataset (i.e., CoGnition) show that our method reduces compositional generalization error rates by 24\% error reduction. 
In addition, our conceptually simple method gives consistently better results than the Transformer baseline on a range of general MT datasets.
\end{abstract}

\section{Introduction}

Neural machine translation (NMT) has achieved competitive performance on benchmark datasets such as WMT \cite{VaswaniSPUJGKP17,Edunov:emnlp18,icml/SoLL19}.
However, the generalizaiton to low-resource domains \cite{emnlp/BapnaF19,emnlp/ZengLSGLYL19,naacl/BapnaF19,iclr/KhandelwalFJZL21} and robustness to slight input perturbations \cite{BelinkovB18,naacl/XuADWJ21} are relatively low for NMT models.
In addition, recent studies show that NMT systems are vulnerable to \textit{compositional generalization} \cite{Lake:icml18,Raunak,Guo:emnlp2020,Li:ACL2021,Dankers2021,Chaabouni2021}, 
namely the ability to understand and produce a potentially infinite (formally exponential to the input size) number of novel combinations of known atoms \cite{chomsky1957,montague1974d,books/el/97/JanssenP97,Lake:icml18,Keysers:iclr2020}.

\begin{figure}[t]
\centering
\includegraphics[width=0.9\linewidth]{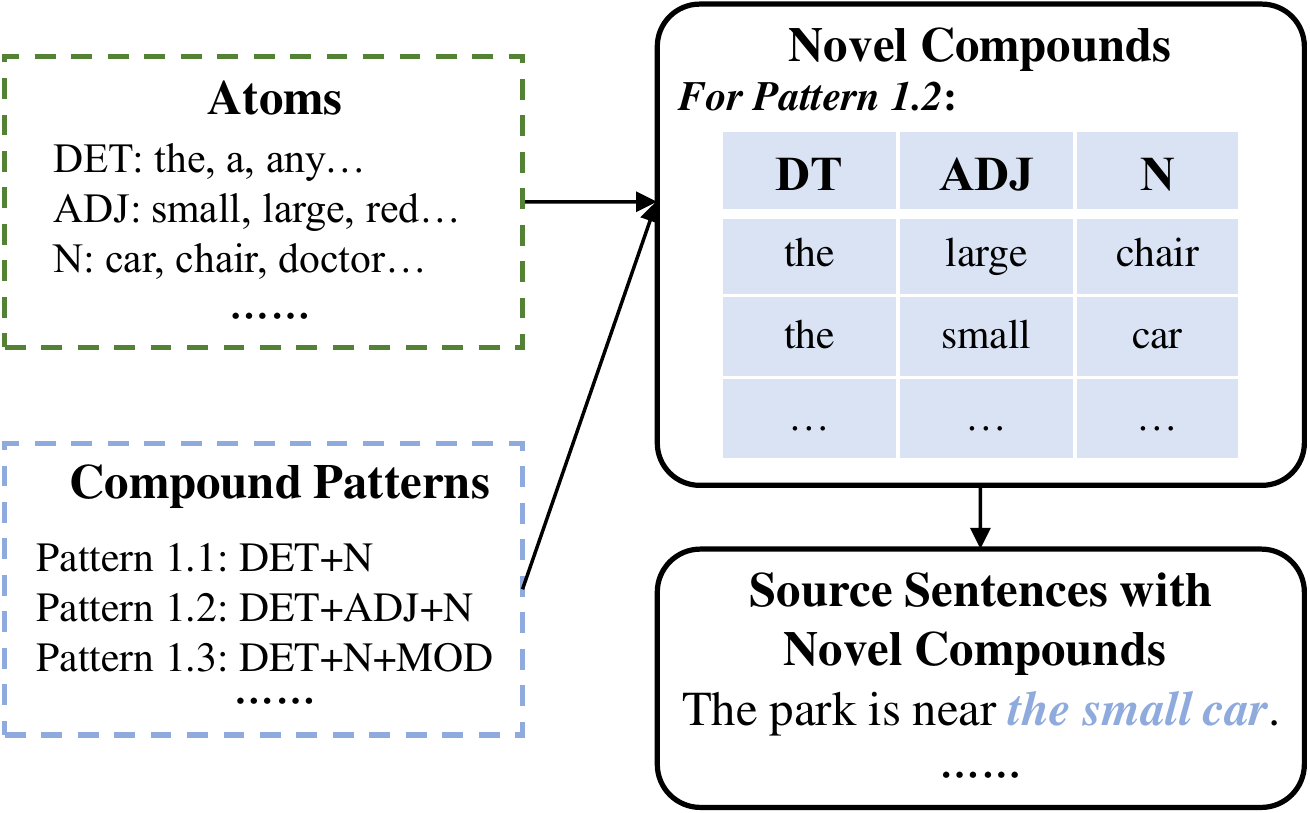}
\caption{
\label{intro}
The novel compounds in the CoGnition test set are constructed by composing a few basis semantic atoms (e.g., determiners (DET), nouns (N), and adjectives (ADJ)) according to the composition patterns. 
The compounds are then put into corresponding source contexts extracted from the training data.
}
\end{figure}

Take CoGnition \cite{Li:ACL2021}, a dedicated MT dataset, for example (Figure \ref{intro}).
Despite that certain instances of translation atoms (e.g., \textit{small}, \textit{large}, \textit{car}, and \textit{chair}) and their semantic compositions (e.g., \textit{small chair} and \textit{large car}) are frequent in training data, unseen compositions of the same atoms (e.g., \textit{large chair}) during testing can suffer from large translation error rates.
Compositionality is also a fundamental issue in language understanding and motivated for translation \cite{books/el/97/JanssenP97,journals/tcs/Janssen98}, which has been 
suggested as being essential for robust translation \cite{Raunak,Li:ACL2021} and efficient low-resource learning \cite{Chaabouni2021}.



The current dominant method to NMT employs a sequence-to-sequence architecture \cite{nips/SutskeverVL14,VaswaniSPUJGKP17}, where an encoder is used to find representations of each input token that thoroughly integrates its sequence-level context information, and a decoder refers to such contextualized representations for generating a translation sequence.
A key reason of failure on compositional generalization is that the correspondence between pairs of {\it token sequences} is modeled as a whole.
Specifically, NMT models are trained end-to-end over large parallel data without disentangling the representation of individual words or phrases from that of whole token sequences.
At the sequence level, the source input sample space is highly sparse mainly due to semantic composition, and small changes to a sentence can lead to out-of-distribution issues \cite{icml/SagawaRKL20,meta21,icml/LiuHCRKSLF21}.

Intuitively, one way to solve this problem is to decouple token-level information from the source sequence by injecting token-level translation distribution
(e.g., $P(petit|small)$) into the source representation.
Given the fact that the source-side contextualized representations encode rich token-level translation information \cite{iclr/Kasai0PCS21,naacl/XuGLX21},
we categorize sparse token-level contextualized source representations into a few representative prototypes over training instances, and make use of them to enrich source encoding. 
In this way, when encoding a sequence, the model observes less sparse prototypes of each token, thus alleviating excessively memorizing the sequence-level information. 

\begin{figure*}[t]
\centering
\includegraphics[width=0.95\linewidth]{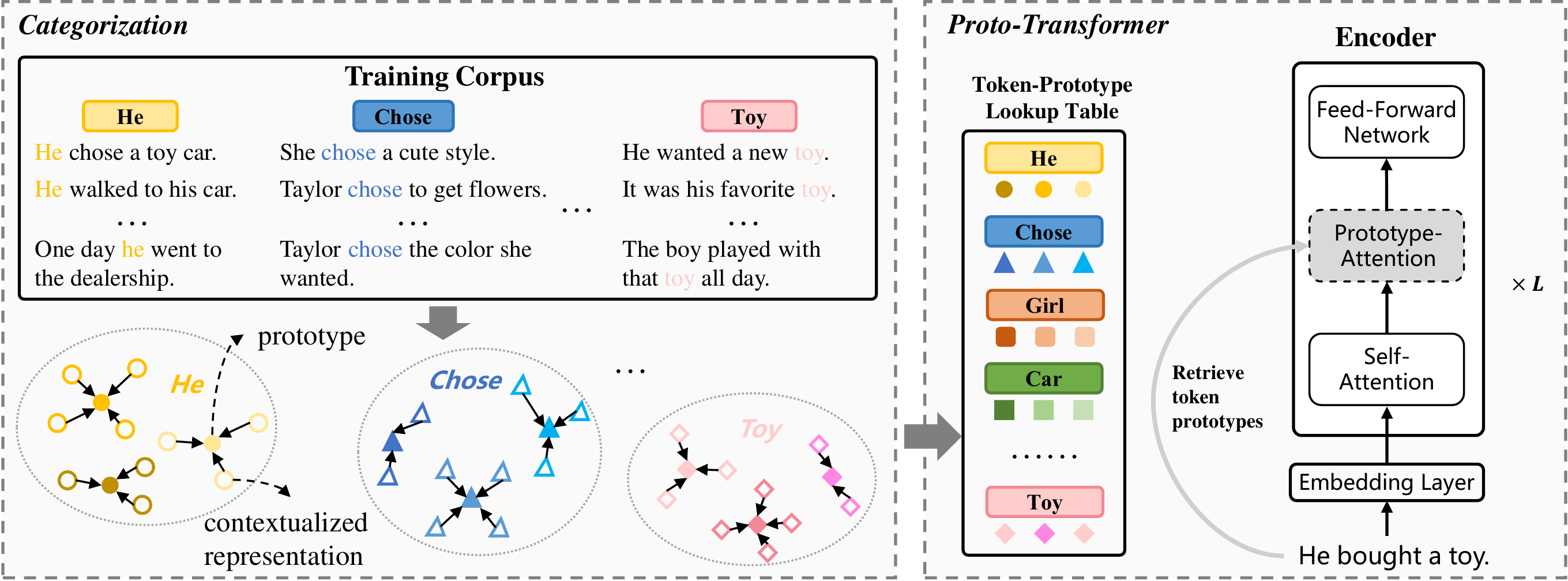}
\caption{
\label{model}
Architecture of Proto-Transformer. The dotted box denotes the prototype-attention introduced in stage 2.}
\end{figure*}

We propose a two-stage framework to train prototype-based Transformer models (Proto-Transformer).
In the first stage, we warm up an initial Transformer model which can generate reasonable representations.
In the second stage, for each token, we run the trained model to extract all contextualized representations over the training corpus.
Then, we perform clustering (e.g., K-Means) to obtain the prototype representations for each token.
Take Figure \ref{model} as an example, for the token ``Toy'', we collect all the contextualized representations and cluster them into 3 prototypes. 
Finally, we extend the base model by fusing the prototype information back into the encoding process through a prototype-attention module, and continue to train the whole model until convergence.

Experimental results on CoGnition show that our method significantly improves novel composition translation by over 24\% error reduction,
demonstrating the effectiveness for tackling the compositional generalization problem.
To further verify the effectiveness on more datasets, we conduct experiments on 10 commonly used MT benchmarks and our method gives consistent BLEU improvement. 
We also present empirical analysis for prototypes and quantitative analysis on compositional generalizaiton.
The comparison between the one-pass and the two-pass training procedure shows that the one-pass method is both faster and more accurate than the two-pass one, demonstrating that more generalizable prototypes extracted from early training phrase are more beneficial to compositional generalization.
Additionally, quantitative analysis demonstrates that our proposed model is better at handling longer compounds and more difficult composition patterns. 
The code is publicly available at https://github.com/ARIES-LM/CatMT4CG.git.

\section{Related Work}

\paragraph{Compositional Generalization}
Recent work \cite{Lake:icml18, cfq} has demonstrated weak compositionality of neural models using dedicated datasets.
Various methods haven been proposed to solve the issue of compositional generalization such as encoding more inductive bias \cite{primitive19,Transcoding19,Realization19,stack20,permutation20,span20}, meta-learning \cite{meta19,meta21}, and data augmentation \cite{good20,recombine21}.
Recently, \citet{Onta/corr/2021} and \citet{emnlp/CsordasIS21} show that the Transformer architecture can perform better on compositional generalization with some modifications.
Although these methods have demonstrated better generalization or interpretability, most of them are limited small vocabulary and limited samples semantic parsing datasets.
In the context of machine translation, \citet{Lake:icml18} construct a small dataset where the training data contains a word \textit{daxy} along with its parallel sentences of a single pattern (e.g., \textit{ I am daxy}, \textit{je suis daxist}) while the test set contains novel patterns (e.g., \textit{He is daxy}). 
However, the experiment is limited in that the test set only consists of 8 samples.
Different from existing work, \citet{Li:ACL2021} propose a large dataset (CoGnition) and construct a large-scale test set that contains newly constructed constituents as novel compounds, so that generalization ability can be evaluated directly based on compound translation error rate.
We proposed a method enhancing compositional generalization on the dedicated dataset of \citet{Li:ACL2021}, while at the same time gives improvements to the machine translation quality at practical test cases.

\paragraph{Neural Machine Translation}
Recent research on NMT has paid increasing attention to robustness \cite{acl/LiuTMCZ18,acl/ChengJME20,naacl/XuADWJ21}, domain adaptation \cite{emnlp/BapnaF19,emnlp/ZengLSGLYL19,naacl/BapnaF19,iclr/KhandelwalFJZL21}, 
and compositional generalization \cite{Lake:icml18,Raunak,Fadaee:acl2020,Guo:emnlp2020,Li:ACL2021}.
\citet{Lake:icml18} propose a simple toy experiment to first show the problem of compositionality.
\citet{Raunak} find that NMT models behave poorly on recombining known parts and generalizing on samples beyond the observed length during training, \citet{Fadaee:acl2020} find that NMT models are vulnerable to modifications such as removal of ad-verbs and number substitutions.
More recently, \citet{Li:ACL2021} observe significant compositional generalization issues on CoGnition, and \citet{Dankers2021} argue that MT is a suitable testing ground to ask how compositional models trained on natural data are.
Our work is in line with the above methods, but we consider a method to address the issue rather than analyse the problem.
Technically, \citet{Raunak} propose to use bag-of-word regularization to refine encoder and \citet{Guo:emnlp2020} propose sequence-level mixup to create synthetic samples.
Different from them, we propose to enhance models' compositional generalization by categorizing contextualized representations, which turns out more effective.

\section{Method}

\subsection{Transformer Baseline}
Given a sequence of source sentence $X=\{x_1,...,x_T\}$, where $T$ denotes the number of tokens, the Transformer encoder \cite{VaswaniSPUJGKP17} first maps $X$ to embeddings, packing them as a matrix $H^0$, and then takes $H^0$ as input and outputs a contextualized sequence representation $H^L\in\mathbb{R}^{d \times T}$, where $d$ and $L$ denote dimension size and the number of layers respectively. 

\paragraph{Attention.}
Formally, given a set of packed query, key, and value matrices $Q$, $K$, and $V$, the dot product attention mechanism are defined as 
\begin{equation}
\text{Attention}(Q,K,V)={\rm Softmax} (\frac{Q^TK}{\sqrt{d}})V,
\end{equation}
where $d$ is the dimension of the key vector.

A typical extension of the above is multi-head attention (MHA), where multiple linear projections are executed in parallel, and the outputs of all heads are concatenated:
\begin{align}
&\text{MHA}(Q,K,V)=W^O[head_1;...;head_h],\\
&head_i=\text{Attention}(W^Q_iQ,W^K_iK,W^V_iV),
\end{align}
where $W^O$, $W^Q_i$, $W^K_i$, and $W^V_i$ are model parameters.

\paragraph{Layer Structure.}
The Transformer encoder has $L$ identical layers, each of which is composed of two sublayers (i.e., self-attention and feed-forward networks). 
In the $l$-th self-attention layer, the query, key, and value matrices are all the hidden states from the previous layer $H^{l-1}$:
\begin{equation}
H^l_a=\text{MHA}(H^{l-1},H^{l-1},H^{l-1}).
\end{equation}
The feed-forward sublayer is a two-layer transformation with ReLU activation:
\begin{equation}\label{ffn}
H^l=W_2^l\text{ReLU}(W_1^lH^l_a+b_1^l)+b_2^l,
\end{equation}
where $W_1^l$, $b_1^l$, $W_2^l$, and $b_2^l$ are trainable parameters.
The layer normalization and residual connection are omitted for brevity.

\subsection{Proto-Transformer}

Our proposed Proto-Transformer extends the base Transformer by introducing a prototype-attention module on top of the self-attention module in each encoder layer, which aggregates token-level prototype representations for each token.

\subsubsection{Prototype-attention}
Assuming that each token $x_t$ in the input sequence is assigned a set of prototype vectors
packed as a matrix $C^{x_t}$ $\in$ $\mathbb{R}^{d \times k}$ where $k$ is the number of prototypes.
The prototype attention aggregates the global prototype information and refines the contextualized representations for each token by the multi-head attention mechanism:
\begin{equation}
H^l_p=\text{MHA}(H^l_a,C,C),
\end{equation}
where $C$ is all the prototype representations of the sequence (i.e., $\{C^{x_1},C^{x_1},...,C^{x_T}\}$), 
and the localness of prototype attention is implemented by mask mechanisms.
The output $H^l_p$ is fed into the feed-forward network (Eq. \ref{ffn}).

The difference between the Proto-Transformer encoder and the Transformer encoder is illustrated by the dotted box in Figure \ref{model}.
By stacking multiple self-attention layers with prototype-attention, the encoder is able to 
exploit less-sparse prototype representations,
preventing the model from over-memorizing the local context.

\subsubsection{Training}\label{training}

\begin{algorithm}[t]
	\caption{The training procedure of Proto-Transformer}
	\label{Algorithm1}
	\small
    \begin{algorithmic}[0] 
			\STATE {\bfseries Input:} 
			Training set $\mathcal{D}$ = $\{(X,Y)\}$, vocabulary $\mathcal{V}$, NMT model $\theta^{(0)}$,
			prototype-attention modules $\theta_+$,
            number of prototype $k$,
			training epochs of stage one $N$,
			token-representation lookup table $\mathcal{Q}$, token-prototype lookup table $\mathcal{P}$
			\STATE {\bfseries Output:} 
			Trained NMT model $\theta^*$
			\STATE Initialize $\theta^{(0)}$ and $\theta_+$
			\STATE \textbf{Stage 1}
			\FOR{$i=1$ {\bfseries to} $N$}
			\STATE $\theta^{(i)}$ $\leftarrow$ TrainModel($\mathcal{D}$, $\theta^{(i-1)}$)
            \ENDFOR
			\STATE \textbf{Stage 2}
			\STATE {\bf for} $X \in \mathcal{D}$ {\bf do}
			\STATE \ \ \ \ {\bf for} each token $x_i$ in $X$ {\bf do}
			\STATE \ \ \ \ \ \ \ \ Add $H_i$ to the corresponding contextualized representation list $\mathcal{Q}[x_i]$ for token $x_i$
			\STATE \ \ \ \ {\bf end for}
			\STATE {\bf end for}
			\STATE Apply K-Means to each token $v_j \in \mathcal{V}$ to obtain prototype representations $C^{v_j}$ and add $(v_j,C^{v_j})$ to $\mathcal{P}$
			\STATE Introduce prototype-attention modules to the base model: $\theta^{(N)} \leftarrow \theta^{(N)}\cup\theta_+$ 
			\STATE Continue training to convergence: 
			\STATE $\theta^*$ $\leftarrow$TrainModel($\mathcal{D}$, $\theta^{(N)}$,$\mathcal{P}$)
	\end{algorithmic}
\end{algorithm}
Proto-Transformer takes a two-stage training process, which is summarized in Algorithm \ref{Algorithm1}.
Here, $\mathcal{D}$ denotes a training corpus, $\mathcal{V}$ denotes the vocabulary of $\mathcal{D}$, $\theta^{(0)}$ denotes initial parameters of a Transformer model, $\theta_+$ denotes parameters of the prototype-attention module, and $k$ and $N$ denote the number of prototypes and training epochs of stage 1 respectively.

\paragraph{Stage 1.}
We first train a base Transformer model for $N$ epochs until it is able to generate reasonable translations.
Given the training corpus $\mathcal{D}=\{(X,Y)\}$, where $X$ and $Y$ denote a source sentence and target translation, respectively, the model is optimized by minimizing cross-entropy loss.
After training for $N$ epochs we obtain a model $\theta^{(N)}$, which has acquired some translation knowledge.

\paragraph{Stage 2.}
We cluster the contextualized representations of each token to obtain its prototype representations. 
In particular, using the model $\theta^{(N)}$, we build a token-representation lookup table $\mathcal{Q}$.
We iterate through the whole training corpus and calculate contextualized representations $\{H_1^{v_i},...,H_{R(v_i)}^{v_i}\}$ of each source token $v_i$ in the vocabulary except punctuation, where $R(v_i)$ is the number of contextualized representations of token $v_i$.
Here we omit the superscript $L$ for simplicity.
Next, for each token $v_i$, we use K-Means \cite{Lloyd82} \footnote{https://scikit-learn.org/stable/modules/classes.html} to cluster the contextualized representations due to its efficiency for large number of samples in high dimensions:
\begin{equation}
    C^{v_i}=\text{K-Means}(H_1^{v_i},...,H_{R(v_i)}^{v_i}).
\end{equation}
All token-prototype pairs are saved in a lookup table $\mathcal{P}$.
Finally, we introduce the prototype-attention modules with parameters $\theta^+$ to the Transformer encoder, and for each source sentence we retrieve the prototype representations from table $\mathcal{P}$.
We continue to optimize the whole parameter set $\{\theta^{(N)}\cup\theta^+\}$ to obtain a final model $\theta^*$. 

As mentioned earlier in the introduction, we compare the above one-pass method to a two-pass method, where we train the base model to convergence instead of for $N$ iterations, before training a Proto-Transformer from scratch which uses clustered prototypes from the base model.
The relative advantage is that the prototypes are taken from a fully trained NMT model, but the disadvantage is that the training time doubles.
In addition, it remains a empirical question whether fully-trained NMT models give prototypes that are more suitable for guiding a final model.
We discuss this in the Section \ref{twopass}.

It is worth noting that a limitation of K-Means is the pre-specification of cluster numbers, which can be different for different tokens in practice.
Non-parametric clustering algorithms such as DBSCAN \cite{dbscan} can potentially solve the problem. However, their runtime complexity is typically $O(n^2)$ or worse, which prevents us from choosing them for large-scale NMT data.


\section{Experiments}

\subsection{Experimental Settings}

\paragraph{Datasets.}
We use CoGnition \cite{Li:ACL2021} to systematically evaluate compositional generalization in MT scenarios, an English$\rightarrow$Chinese (En$\rightarrow$Zh) translation dataset.
In consists of a training set of 196,246 sentence pairs, a valid set and a test set of 10,000 samples. 
In particular, it has a dedicated test set (i.e., CG-test set) consisting of 10,800 sentences containing novel compounds, so that the model's ability of compositional generalization can be measured by the ratio of compounds that are correctly translated. 
In addition, we choose 9 machine translation tasks from IWSLT, WMT and JRC-Acquis to verify the general effectiveness of our methods.
The dataset statistics are shown in Appendix \ref{appd_data}.

\begin{table*}[t]
\centering
\small
\begin{tabular}{c||c|c|c|c}
\hline
\multirow{2}{*}{\bf Model} & \multicolumn{4}{c}{\bf Compound Translation Error Rate (CTER) $\downarrow$} \\
\cline{2-5}
& NP & VP & PP & Total \\
\hline
\hline
Transformer & 24.74\%/55.16\% & 24.82\%/59.54\% & 35.71\%/73.94\% & 28.42\%/62.88\% \\
\hline
Transformer-Deep & 25.11\%/54.11\% & 28.14\%/60.43\% & 37.38\%/75.24\% & 30.21\%/63.26\% \\
\hline
Transformer-Small & 22.14\%/47.21\%  & 23.55\%/53.60\% & 32.02\%/69.64\% & 25.91\%/56.82\% \\
\hline
Bow & 22.16\%47.89\% & 24.83\%/55.57\% & 	35.04\%/73.21\% & 27.34\%/58.89\% \\
\hline
SeqMix & 24.52\%/49.71\% & 26.88\%/58.90\% & 34.36\%/73.09\% & 28.59\%/60.57\% \\
\hline
Transformer-Rela & 22.69\%/51.20\% & 24.89\%/57.17\% & 34.94\%/71.73\% & 27.50\%/60.03\% \\
\hline 
\hline
Proto-Transformer & \bf 14.07\%/36.45\%  & \bf 22.13\%/50.90\% & \bf 28.85\%/68.15\% & \bf 21.69\%/51.84\% \\
\hline
\end{tabular}
\caption{
Compound translation error rate (CTER) on CoGnition.
We report instance-level and aggregate-level CTERs, separated by ``/''. NP, VP, and PP denote noun phrases, verb phrases and positional phrases, respectively, three compound types in the compositional generalization test set (CG-test set).
}\label{mainexp}
\end{table*}

\paragraph{Setup.}
\label{setup}
We use Transformer \cite{VaswaniSPUJGKP17} as our baseline models implemented using the Fairseq toolkit \cite{ott2019fairseq}.
For CoGnition and IWSLT, we use the Transformer \textcolor{gray}{iwslt\_de\_en} setting
while for the others
we use the \textcolor{gray}{transformer\_base} setting.
Following previous work, for IWSLT and JRC-Acquis En$\leftrightarrow$Es, we use beam search with width 5 and length penalty 0.6 for inference, whereas for the other datasets we set the beam width as 4.
For CoGnition, $N$ and $k$ are set as 8 and 3 based on the validation set, and $N$ and $k$ for the other datasets are shown in Appendix \ref{appd_data}. 
Empirically, $N$ is recommended to chosen from 10\% to 25\% of the total training epochs.
We conduct categorization for all tokens except punctuation and low-frequent words, and the token on CoGniton is word and is subword on other datasets
For reproducibility and stability, we train 6 models with seeds provided in \cite{Li:ACL2021} for each method and report the average performance.

\paragraph{Evaluation Metrics.}
We use compound translation error rate (CTER; \cite{Li:ACL2021}) to measure model performance on CoGnition.
Specifically, \textit{instance-level} CTER denotes the ratio of samples where the novel compounds are translated incorrectly, while \textit{aggregate-level} CTER denotes the ratio of compounds that suffer at least one incorrect translation in corresponding contexts.
To calculate CTER, \citet{Li:ACL2021} provide a manually collected dictionary for all the atoms based on the training set, since each word may have different translations. 
We also conduct human evaluation (Appendix \ref{appd_human}) as a supplement.
We conduct evaluation using BLEU \cite{bleu} for the other datasets. 


\subsection{Baseline Methods}
We compare our method with following baselines: (1) Transformer \cite{VaswaniSPUJGKP17}, which uses the same settings as \citet{Li:ACL2021};
(2) Transformer-Small, a more compact Transformer model with 4 layers and 256 hidden size;
(3) Transformer-Deep, which increases the number of encoder layers to 8 to take the same parameters as Proto-Transformer;
(4) Transformer-Rela, which replaces absolute positional encoding with a relative one, an important component for compositional generalizaiton demonstrated in \cite{emnlp/CsordasIS21,Onta/corr/2021};
(5) Bow \cite{Raunak}: which uses bag-of-words loss to regularize the encoder based on the observation that the encoder representations are much weaker on unseen composition; 
and (6) SeqMix \cite{Guo:emnlp2020}: which synthesizes examples by interpolating embeddings\footnote{The performance is the best when the hyper-parameters of Beta distribution are set to 1.0, which is consistent with \citep{Guo:emnlp2020}.}.

\subsection{Results on CoGnition} \label{result_cog}

The main results on CoGnition are shown in Table \ref{mainexp}.
Transformer gives instance-level and aggregate-level CTERs of 28.42\% and 62.88\%, respectively. 
In comparison, Proto-Transformer gives a score of 21.69\% and 51.84\%, respectively, with a significant improvement of 6.73\% and 11.04\% accordingly.
Moreover, Proto-Transformer outperforms all baseline systems significantly, indicating that categorization on the contextualized representations is more beneficial to compositional generalization.
We also calculate the BLEU scores, and Proto-Transformer and Transformer obtain 60.1 and 59.5, respectively. 
The BLEU scores are all relatively high since the sentences on the CG test set are similar to the sentences in training data except the novel compounds.
For CoGnition, CTER is more accurate and suitable than BLEU because the translation dictionary processes multiple accurate translations of each compound better.

Since Proto-Transformer brings some extra parameters (approximately 6M), we investigate whether the performance improvement is derived from the increase of model parameters.
As can be seen, Transformer-Deep performs poorly on the CG-test set, indicating that only increasing model capacity may be harmful since it leads to worse over-fitting to sequence-level distributions. 
Besides, the more compact model Transformer-Small yields better results than Transformer and Transformer-Deep but lags far behind Proto-Transformer.
This shows that model size is useful but not sufficient for solving compositional generalization.

Proto-Transformer performs better than Bow, indicating that the encoder representations refined by categorization are more adequate than by the regularization technique.
Compared to SeqMix, the improvement of Proto-Transformer is more significant (2.31\% vs 11.04\% aggregate-level CTER).
SeqMix reduces representation sparsity via linear interpolation in the input embedding space, and we conjecture that the stochastically synthetic samples may be unreasonable and harmful to model training. 
Relative positional embedding \cite{conf/naacl/ShawUV18} is demonstrated to be important for compositional generalization.
Specifically, Transformer-Rela reduces CTERs by 0.92\% and 2.85\% but is inferior to ours, indicating that the prototypes bring more than positional information.


\subsection{Results on General MT Datasets}



\begin{table}
\centering
\small
\begin{tabular}{c|c|c|c}
    \hline
     \multirow{2}{*}{\textbf{Dataset}} & \multirow{2}{*}{\textbf{Direction}}& \multicolumn{2}{c}{\bf BLEU }  \\
     \cline{3-4}
     &&\textbf{Transformer} & \textbf{Proto-TF}  \\
     \hline
     \hline
    \multirow{4}{*}{\bf IWSLT} &  En$\Rightarrow$De &  28.44 & \textbf{28.96} \\
\cline{2-4}
& En$\Rightarrow$It &   28.24 & \textbf{28.87}   \\
\cline{2-4}
& En$\Rightarrow$Vi &   30.19 & \textbf{30.96}   \\
 \hline
\multirow{3}{*}{\bf WMT} & En$\Rightarrow$Ro &  32.46 & \textbf{33.37} \\
\cline{2-4}
& Ro$\Rightarrow$En &  32.49 & \textbf{33.27}   \\
\cline{2-4}
& En$\Rightarrow$Fi &  21.39 & \textbf{22.11}   \\
\cline{2-4}
& En$\Rightarrow$De &  27.95 & \textbf{28.49}   \\
\hline
{\bf JRC-} & En$\Rightarrow$Es &  60.90/60.32 & \textbf{61.56/60.90} \\
\cline{2-4}
\textbf{Acquis}& Es$\Rightarrow$En &   63.21/62.59 & \textbf{63.73/63.10} \\
\hline
\end{tabular}
    \caption{BLEU on commonly used MT datasets.
    For JRC-Acquis, we follow previous work to report results on both the valid and test sets, separated by ``/''. The performance of our model is significantly better than Transformer ($p<0.05$) \cite{sigtest}.}
    \label{tab:results_general}
\end{table}



We conduct experiments on several general MT benchmarks, where the translation compound error rate (CTER) cannot be calculated directly.
The performances on IWSLT, WMT and JRC-Acquis are presented in Table \ref{tab:results_general}. 
Compared with Transformer, our model achieves consistent improvement (0.67 BLEU), demonstrating its effectiveness under general evaluation settings. 
Proto-Transformer outperforms Transformer by 0.91 and 0.78 BLEU scores on WMT'16 En$\rightarrow$Ro and Ro$\rightarrow$En respectively.
For JRC-Acquis, our model achieves an average improvement of 0.62 BLEU score on En$\rightarrow$Es and 0.56 BLEU score on Es$\rightarrow$En.
On the largest dataset WMT'16 En$\rightarrow$De, our model also performs better than Transformer by 0.54 BLEU score.
In addition, the large datasets possibly benefit from the additional parameters (about 6M).
We run an experiment of a model with 8 encoder layers on EN-DE, and it achieves 28.02 BLEU, inferior to our model with same number of parameters.

Note that the BLEU scores in the general datasets can reflect translation quality, but has several limitations in the measurement of generalization or robustness. 
For example, when a sentence is almost translated correctly but has few serious errors, it can achieve a high BLEU score but suffers severe semantic distortion.
For instance, as shown in \cite{Li:ACL2021}, the model mis-translates ``He became sick from eating all of the peanut butter on the ball`` into ``He became sick from eating all of the peanut butter on the field``.
With a minor mistake on the compound ``on the ball``, the model achieves a sentence-level BLEU of 61.4, but the full sentence meaning is largely affected.

\section{Analysis}

\subsection{Prototypes}

\paragraph{Effect of the Number of Prototypes.}

\begin{table}[t]
\centering
\small
\begin{tabular}{c|c|c|c}
\hline
\bf $k$ & \bf CTER & \bf $k$ & \bf CTER \\
\hline
\hline
0  & 28.42\%/62.88\% & 3  & 21.69\%/51.84\% \\
1  & 23.60\%/54.33\% & 4  & 23.04\%/52.70\% \\
2  & 22.58\%/53.19\% & 5  & 24.98\%/54.71\% \\
\hline
\end{tabular}
\caption{
CTERs against different number of prototypes on the CG-test set. ``0'' denotes to the Transformer baseline.
}\label{anak}
\end{table}

The number of prototypes $k$ controls the granularity of context used to refine the representation learning, and is determined it based on the model loss on the development set.
In this experiment, we investigate its influence on generalization performance.
As shown in Table \ref{anak}, incorporating prototypes can reduce the translation error caused by unseen compounds, and the model obtains the best generalization performance when $k$ is 3. 
Intuitively, using too many prototypes dilutes the concentration effect, leading to overfitting again, while too few prototypes limits the expressiveness for polysemous words, e.g., one prototype represents a single sense. 

\begin{table}[t]
\centering
\small
\begin{CJK*}{UTF8}{gbsn}
\begin{tabular}{c|c|c|c}
\hline
Tgt tokens & PT & TF & Src tokens \\
\hline
\hline
yesterday & 1 &	3 &	\textbf{ieri},psd,victor \\
\hline 
limited	& 2	& 5	& \textbf{limitata},\textbf{limitat},comert,apel, ... \\
\hline
six	& 1	& 6	& \textbf{sase},procente,saptamani,din, ... \\
\hline
agriculture	& 1	& 4	& \textbf{agricultura},plasa, ... \\
\hline
culture	& 2	& 5	& \textbf{cultura},\textbf{culturii}, profesion... \\
\hline
republic &	1 &	3 &	\textbf{republica},liderului, ... \\
\hline
simply &	1 &	3 &	\textbf{simplu},nu,individuala, ... \\
\hline
november &	1 &	6 &	\textbf{noiembrie},propaganda,2008, ... \\
\hline
tomorrow &	1 &	3 &	\textbf{maine},dimineata,amiaza, ... \\
\hline
saturday &	1 &	3 &	\textbf{sambata},seara,dimineata, ... \\
\hline
\end{tabular}
\caption{
\label{case_align}
Word translation variations induced from Transformer (TF) and Proto-Transformer (PT). 
The columns PT and TF denote the number of aligned Romanian tokens, and
the bold characters display the tokens induced from Proto-Transformer.
}
\end{CJK*}
\end{table}

\paragraph{Effect on Token-level Translation Consistency.}
We hypothesize that token alignments induced by Proto-Transformer are less ambiguous for tokens having settled meaning.
To verify it, we approximately measure the number of distinct translations for 10 selected English tokens in the WMT Ro$\rightarrow$En dataset, by counting the number of Romanian tokens that are aligned to the English tokens in the test set with the method of \citet{emnlp/ChenLCJL20}.
As can be seen in Table \ref{case_align}, Proto-Transformer attends to much fewer token types than Transformer. 
This shows that Proto-Transformer's translations are less prone to context changes, which indicates the sparsity reduction of input sample space.
The finding also partly explains the improvement on the Ro$\rightarrow$En dataset.
The above observations verify our intuition in the introduction that adding prototypes to the source representation enhances the model's knowledge on token-level translation consistency, which leads to better compositional generalizaiton.

It is noteworthy that increase translation consistency may increase translationese effects \cite{vanmassenhove-etal}. We argue that there can be a trade-off between the consistency modeling and translationese. Our approach does not force the model to generate highly consistent translation, but aims to alleviate the vulnerability to context changes by introducing several prototypes.

\begin{figure}[t]
\centering
\includegraphics[width=1.0\linewidth]{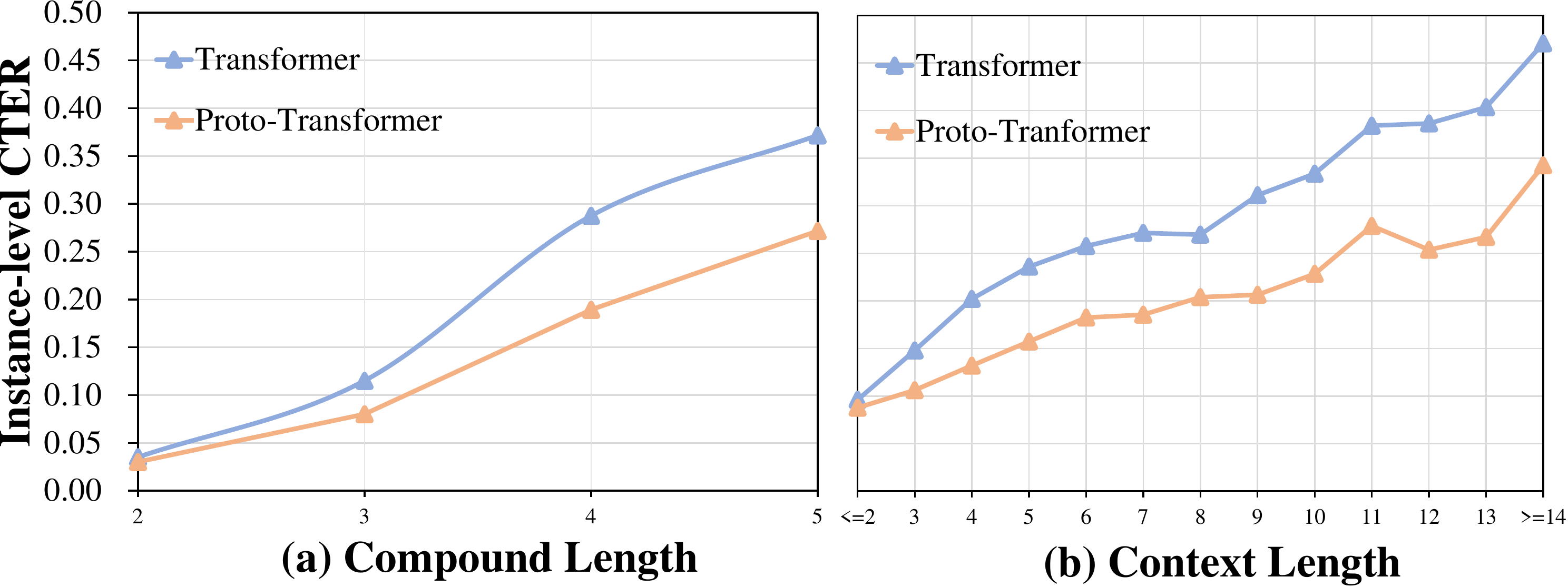}
\caption{
\label{comp-length}
CTERs of Proto-Transformer and Transformer over different compound and context lengths.
}
\end{figure}

\paragraph{One-pass vs Two-pass.}\label{twopass}
The possible advantage of the two-pass training procedure (Two-Pass) mentioned in Introduction is that the quality of the prototype representations generated by a converged model may be better.
Specifically, Two-Pass achieves 24.36\% instance-level and 55.69\% aggregate-level CTER, outperforming Transformer by 4.06\% and 7.19\%, respectively, but is noticeably inferior to Proto-Transformer (21.69\% and 51.84\%).
This observation may relate to the observation that neural models learn more generalizable features in the early phase before memorization \cite{icml/Arpit17,iclr/StephensonPGHTC21}.
The one-pass approach can not only reduce the training cost (11,103 seconds vs 6,286 seconds) but also leverage more generalizable features to better improve models' compositional generalization. 
In addition, the prototypes can be initialized randomly and trained along with the model, and this variant obtains 26.24\% and 57.94\% CTERs.
It preforms much worse than Proto-Transformer (21.69\% and 51.84\%) though they have identical number of parameters and architecture, indicating the effectiveness of explicit categorization during training.

\subsection{Effects on Compositional Generalization}


\paragraph{Composition Length.}

Longer compounds are harder to generalize as they contain richer semantic information \cite{Li:ACL2021}.
We classify the test samples by compound length and context length, and calculate the instance-level CTER. 
In Figure \ref{comp-length}, we can observe that the advantage of Proto-Transformer grows larger in generalizing longer compounds and context. 
In particular, Proto-Transformer gives a lower CTER by 12.80\% over samples with context longer than 13 tokens.
The underlying reason can be that longer compounds or contexts are more sparse in input space,
and Proto-Transformer alleviates sparseness by putting token distributions into representations via prototypes.

\begin{table*}[!ht]
\centering
\small
\begin{CJK*}{UTF8}{gbsn}
\begin{tabular}{p{4.7cm}<{\centering}|p{4.7cm}<{\centering}|p{5.4cm}<{\centering}}
\hline
\hline
\textbf{Source} & \textbf{Transformer} & \textbf{Proto-Transformer} \\
\hline
\hline
& 然而，使她沮丧的是， & 但令她沮丧的是，\\ 
Yet to her dismay something terrible & 一个可怕的男友醒来了。 & 某种可怕的东西把那个傻男友吵醒了。\\
\textbf{woke the silly boyfriend up}.  & (Yet to her dismay  & (Yet to her dismay something terrible \\
& \textbf{a silly boyfriend woke up}.) &  \textbf{woke the silly boyfriend up}.) \\
\hline
& 每天早上6点，& 每天早上6点，一只鸟 \\ 
Every morning , a bird \textbf{woke the silly}& 一只愚蠢的鸟叫醒去上班。 & 叫醒那个愚蠢的男朋友去上班。\\
\textbf{boyfriend up} for work at 6 am.  & (every morning , \textbf{a silly bird woke up}  & (Every morning , a bird \textbf{woke the silly}\\
&  for work at 6 am.) &   \textbf{boyfriend up} for work at 6 am.) \\
\hline
His dog \textbf{woke the silly boyfriend up} & 他的狗半夜醒来了那个傻男友。& 他的狗半夜把那个傻男友吵醒了。  \\ 
in the middle of the night. & (His dog \textbf{woke up} in the middle of & (His dog \textbf{woke the silly boyfriend up} \\
&  the night, \textbf{the silly boyfriend}.) & in the middle of the night.) \\
\hline
\hline
\end{tabular}
\caption{
\label{case}
Example translations.
The bold characters denoting the novel compounds and corresponding translations.
}
\end{CJK*}
\end{table*}

\begin{figure}[t]
\centering
\includegraphics[width=0.8\linewidth]{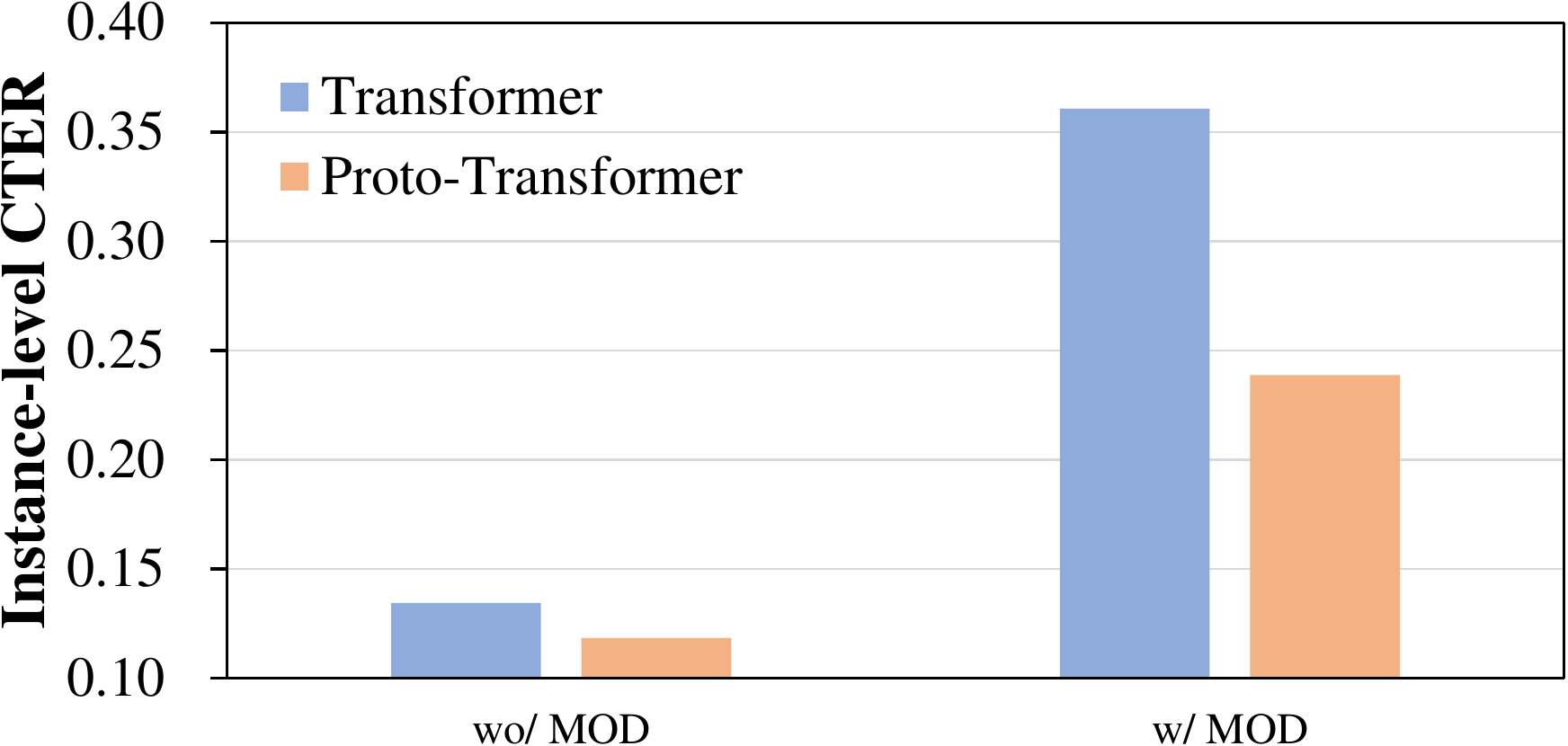}
\caption{
\label{comp-type}
CTERs 
on compounds w/o and w/ MOD.
}
\end{figure}

\begin{figure}[t]
\centering
\includegraphics[width=0.7\linewidth]{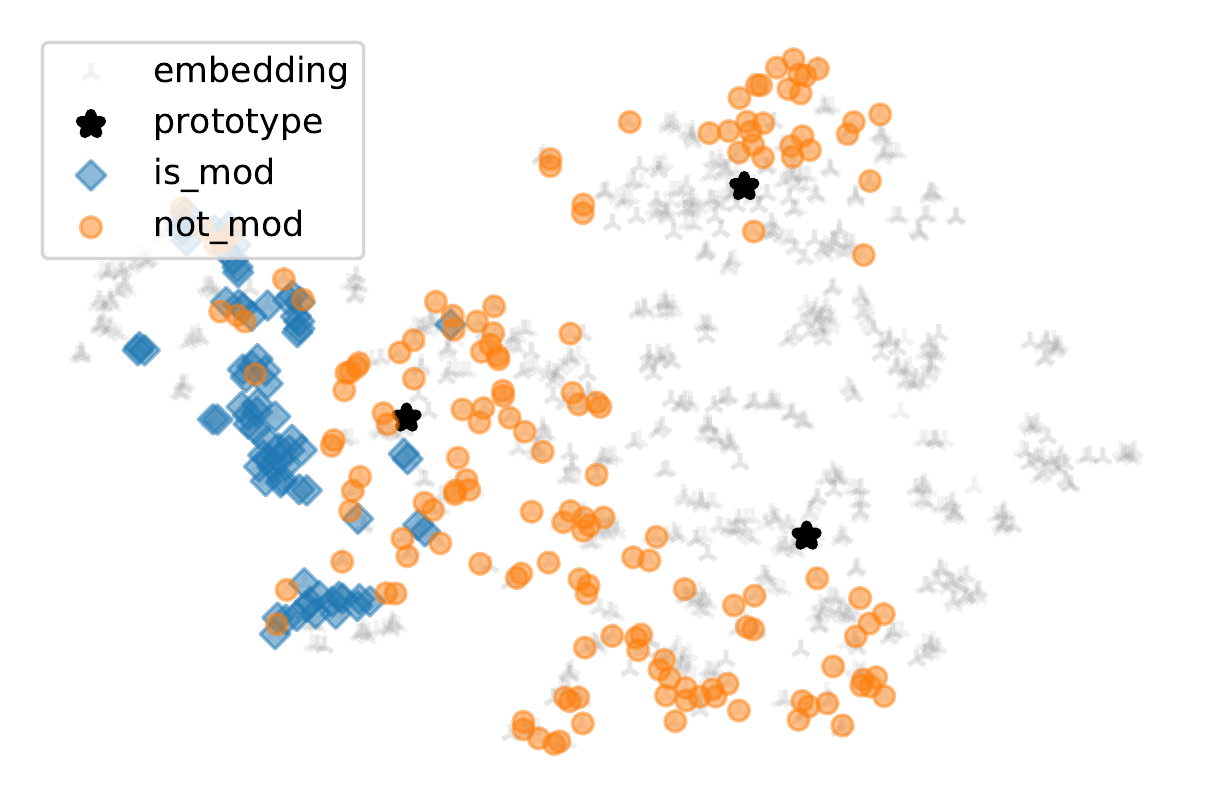}
\caption{
\label{cluster-visual}
Visualization of contextualized embeddings (grey tri-down markers) and 3 prototypes (black stars) for the token \textit{liked}. 
The blue diamonds and orange circles denote the case where \textit{liked} appear in the context \textit{he liked}, and the former one denotes the case where \textit{he liked} serves as MOD.
}
\end{figure}

\paragraph{Modifier.}
One challenging type of novel compounds in CoGnition is the postpositive modifier atom (MOD), which is constructed to enrich the information of its preceding word (e.g., \textit{he liked} in the sentence \textit{he bought the car he liked}). 
The difficulty of translating compounds with MOD lies in word reordering from English to Chinese. 
We divide the test samples into two groups according to compounds with or without MOD (Figure \ref{comp-type}). Proto-Transformer demonstrates larger advantage over Transformer in translating the compounds with MOD, showing its superiority in processing complex semantic composition.
To further understand the effectiveness, we choose the token \textit{liked}, a core part of the MOD atom \textit{he liked}, and visualize its representations with t-SNE \cite{t-SNE}.
In Figure \ref{cluster-visual}, 
we can see that the representations of \textit{he liked} serving as MOD are concentrated at the leftmost prototype, while the representations of \textit{he liked} not serving as MOD scatter on the other prototypes. 
Through representation categorization, Proto-Transformer finds a specialized prototype which abstracts the knowledge of \textit{liked} serving as a part of MOD, and
refers to the MOD prototype when processing compounds with MOD.


\paragraph{Case Study}\label{appd_case}
We present 3 source samples containing a novel compound \textit{woke the silly boyfriend up} and 5 atoms, i.e., \textit{woke}, \textit{the}, \textit{silly}, \textit{boyfriend} and \textit{up}, and their translations in Table \ref{case}.
For all samples, correct translations should contain complete semantic meaning of the compound.
Proto-Transformer correctly translates the compound along with different contexts across all samples, while Transformer suffers various mistakes.
In the first sample, the translation of \textit{something terrible} is omitted by Transformer. 
In the second sample, Transformer omits 
\textit{silly boyfriend} and mistranslates \textit{woke the silly boyfriend up} into \textit{a silly bird woke up}.
Similar mistake can be observed in the third example, where Transformer mistranslates \textit{his dog woke the silly boyfriend up} into \textit{his dog woke up}.
Transformer overfits contexts and makes errors on unseen compositions, while our model is more stable.

\section{Conclusion}
We investigated a conceptually simple method for
enhancing compositional generalizaiton of NMT models, proposing a two-stage training framework to fuse prototype representations into the encoding process of Transformer.
Experiments on CoGnition show the effectiveness of our method on compositional generalization, and extensive results over 9 translation tasks verify the generality of our method.
To our knowledge, we are the first to propose token-level categorization for NMT, achieving promising performance on both a large-scale compositional generalization dataset and general datasets.

\section*{Acknowledgements}
The work is funded by the Zhejiang Province Key Project 2022SDXHDX0003.
We would like to thank all of the anonymous reviewers and Jiali Zeng for the helpful comments.
Yue Zhang is the corresponding author.

\bibliography{anthology,acl_latex}
\bibliographystyle{acl_natbib}

\appendix

\newpage

\section{Human evaluation}\label{appd_human}

\begin{table}[!h]
\centering
\begin{tabular}{c|c|c}
\hline
 Model & Adequacy & Fluency \\
\hline
Transformer & 4.32 & 4.27 \\
Proto-Transformer & 4.51 & 4.54 \\
\hline
\end{tabular}
\caption{Human evaluation on adequacy and fluency.}
\label{Human_evaluation}
\end{table}

We conduct the human evaluation for translations in terms of adequacy and fluency. 
We randomly sample 100 sentences from the CG-test set of CoGnition, and invite three annotators to evaluate the translation adequacy and fluency ranging from one to five.
The five point scale for adequacy indicates how much of the meaning expressed in the reference translation is also expressed in a hypothesis translation: 5 = All, 4 = Most, 3 = Much, 2 = Little, and 1=None.
The five point scale for fluency indicates how fluent the translation is: 5 = Flawless, 4 = Good, 3 = Non-native, 2 = Disfluent, and 1 = Incomprehensible.

The average of the scores from the three annotators is taken as the final score, and the results of the baseline and our model are shown in Table \ref{Human_evaluation}.
Compared with Transformer, Proto-Transformer improves adequacy and fluency by 0.19 and 0.27, respectively. 
Proto-Transformer achieves more accurate translation of the novel compounds by alleviating the problem of compositional generalization, and possibly make the other part of the sentence to be translated better.

\begin{table*}
\centering
\small
\begin{tabular}{lcccccccccc}
\hline
\multirow{2}{*}{} & CoGnition & \multicolumn{2}{c}{IWSLT'14}  & IWSLT'15 & \multicolumn{3}{c}{WMT'16} & WMT'17 & \multicolumn{2}{c}{JRC-Acquis} \\
\hline

& En-Zh & En-De & En-It & En-Vi & En-Ro & En-Ro & En-De & En-Fi & En-Es & Es-En \\
\hline
\#Train & 196k & 157k & 175k  & 133k & 608k & 608k & 4.5M & 2.6M & 679k & 679k \\
\hline
\#Valid & 10k & 7k & 1k &  1.6k & 2k & 2k & 3k  & 9k   & 25k & 25k\\
\hline
\#Test & 10k & 7k & 0.9k & 1.3k & 2k & 2k & 3k   & 3k & 26k & 26k\\
\hline
\ N & 8 & 12 & 10 & 8 & 8 & 11 & 15 & 12 & 8 & 8\\
\hline
\ k & 3 & 4 & 4 & 3  & 4 & 6 & 5 & 6 & 4 & 4 \\
\hline
\end{tabular}
\caption{Dataset statistics and hyper-parameters.}
\label{Dataset_Statistics}
\end{table*}


\section{Dataset Statistics and Hyper-parameters}\label{appd_data}

We list the statistics and the introduced hyper-parameters of all the datasets in Table \ref{Dataset_Statistics}. 

For IWSLT'14 English$\leftrightarrow$German (En$\leftrightarrow$De), IWSLT'14  English$\rightarrow$Italian (En$\rightarrow$It), 
and IWSLT'15 English$\rightarrow$Vietnamese (En$\rightarrow$Vi), we use Moses tokenizer\footnote{https://github.com/moses-smt/} and apply joint BPE \cite{Sennrichbpe} with 10,000 merge operations.
For WMT'16 English$\leftrightarrow$Romanian (En$\leftrightarrow$Ro), we use the processed data from \citet{Lee:emnlp18}.
For WMT'16 English$\rightarrow$German (En$\rightarrow$De) and WMT'17 English$\rightarrow$Finnish (En$\rightarrow$Fi), we apply joint BPE with 37,000 and 32,000 merge operations, respectively. 
For JRC-Acquis English$\leftrightarrow$Spanish (En$\leftrightarrow$Es) we use the datasets processed by \citet{GuWCL:aaai18}. 
For WMT'17 En-Fi, we use the concatenation of newstest2015, newsdev2015, newstest2016 and newstestB2016 as the development set, and the newstest2017 as the test set.

\section{Target-side Prototypes.}
Target-side prototypes possibly contain more bilingual translation knowledge since the decoder processes target sentences based on source representations.
We are interested in whether target-side prototypes can be used to enhance compositional generalization and/or further improve Proto-Transformer.
To answer this question, we first extract prototypes from the decoder and incorporate them back to the decoder using the same mechanism of Proto-Transformer.
The model can reduce CTERs to 25.93\%/58.21\%, largely inferior to using source-side prototypes (21.69\%/51.84\%).
We also try to incorporate target-side prototypes to the decoder based on Proto-Transformer but it gives no noticeable improvement, achieving 21.65\% and 51.71\% CTERs, respectively.
The underlying reason is connected a recent finding that translation already happens in the source encoding and the representations from encoders contain sufficient translation knowledge \cite{iclr/Kasai0PCS21,naacl/XuGLX21}.

\section{Computational Cost}
Our framework adds some computational overheads including extracting contextualized representations and conducting the clustering algorithm.
The former only requires a single forward pass over the training set, merely amounting to a fraction of the cost of training for one epoch. 
Thanks to pytorch implementation of K-means algorithms which utilize GPU for faster matrix computations, the clustering is friendly for the large-scale datasets and it can be much faster with parallelization and more powerful hardware.
Since the number of prototypes is a constant $k$, the complexity of prototype-attention is $O(kT)$, linear with respect to sequence length $T$.

\end{document}